\documentclass[journal]{vgtc}

\vgtccategory{Application/Design Study}
\manuscriptnote{}

\title{DECODE: Diagnosing Interpretability Failures in \\ Automated Interpretation Process}

\author{
  \authororcid{Dohyun Kim}{0009-0004-7200-9079},
  \authororcid{Jaeung Lee}{0009-0004-7718-5653},
  \authororcid{Donghee Hong}{0009-0005-0890-2556},
  and \authororcid{Jaemin Jo}{0000-0002-5207-6010}
}

\authorfooter{
  \item Dohyun Kim is with KAIST, Seoul, Republic of Korea. This work was done while at Sungkyunkwan University. E-mail: kimdoh0423@kaist.ac.kr.
  \item Jaeung Lee is with Samsung Research, Seoul, Republic of Korea. E-mail: dlwodnd00@gmail.com.
  \item Donghee Hong is with University of Maryland, College Park, Maryland, United States. E-mail: donghee@umd.edu.
  \item Jaemin Jo is with Sungkyunkwan University, Suwon, Republic of Korea, and is the corresponding author. E-mail: jmjo@skku.edu.
}

\abstract{%
Tools for interpreting the internal representations of Large Language Models (LLMs) increasingly rely on LLMs themselves to both generate and evaluate explanations, creating a self-referential automated interpretation process with minimal human oversight.
This automation raises significant reliability concerns, which are exacerbated by the absence of ground truth.
In practice, it is difficult to validate explanations or to attribute interpretability failures to their source—whether they originate from the tool's extracted representations or from the LLM used in the process.
To address these challenges, we present DECODE, a visual analytics system that reintroduces human oversight into the automated interpretation process.
Our approach decomposes the diagnostic workflow into three progressive stages, isolating failure sources ranging from structural defects to explanation-level misalignments.
Drawing inspiration from inter-coder consensus in qualitative analysis, we utilize consensus among independently generated explanations and their evaluation metrics as a proxy for reliability.
To further scale human inspection, we employ a three-phase active seeding protocol that strategically recommends high-priority instances for manual review.
Finally, we demonstrate the effectiveness of our approach through a case study on 16,384 Sparse Autoencoder features, showing that it can efficiently identify and attribute interpretability failures at scale.
}


\graphicspath{{figs/}{./}}
\usepackage{tabu}
\usepackage{booktabs}
\usepackage{mathptmx}
\usepackage{amssymb}
\usepackage{kotex}
\usepackage{makecell}

\usepackage{mathtools}
\usepackage{xspace}
\usepackage{multirow}

\begin{document}

\firstsection{Introduction} 

\maketitle

Large Language Models (LLMs) are increasingly deployed in critical domains, yet their internal decision-making processes remain opaque.
Mechanistic interpretability research has emerged to address this challenge by decomposing internal representations of LLMs into interpretable units \cite{geva2021transformer, bills2023language, olsson2022context, wang2023interpretability, bricken2023monosemanticity}, often referred to as latent features.
At the scale of modern models, this decomposition produces hundreds of thousands of features, far exceeding the capacity of manual inspection. Researchers therefore resort to LLM-based automation, where LLMs both generate explanations of the features and evaluate their faithfulness \cite{bills2023language, paulo2025automatically, gao2025scaling, ameisen2025circuit, camunas2025openmaia, han2025sage}.

While this automation alleviates the scalability bottleneck, it introduces a self-referential validation loop in which LLMs both generate and evaluate explanations with minimal human oversight.
Recent agentic approaches replace single-pass generation with iterative hypothesis testing to improve faithfulness \cite{camunas2025openmaia, han2025sage}, yet the final judgment still rests with an LLM, whose errors may remain undetected.
Findings that LLMs can coordinate to produce deceptive interpretable feature labels \cite{lermen2025deceptive} further underscore the fragility of such a fully automated process.

This self-referential loop poses two compounding challenges. First, learned features lack ground-truth explanations, leaving no external reference for validating LLM-generated outputs.
Second, existing end-to-end metrics conflate structurally distinct failure types into a single score, obscuring where and why an interpretability failure occurs.

We argue that human oversight remains essential for breaking this self-referential loop and grounding the diagnostic process in interpretable and accountable judgment.
To this end, we present DECODE, a visual analytics system that helps interpretability researchers perform an initial diagnosis of the automated interpretation process before downstream use.
To address the absence of ground truth, DECODE uses consensus among independent LLM explanations and their evaluation metrics as a proxy for reliability, drawing on inter-coder consensus in qualitative analysis.
Rather than relying on end-to-end metrics, it decomposes the diagnostic process into three stages: researchers determine whether each feature captures a complete concept, whether its explanations faithfully describe its activation pattern, and where each failure originates, assigning every feature a diagnostic category that guides corrective action.
Semi-automatic labeling through active seeding further scales human judgment beyond direct inspection.

Our contributions are as follows:
\begin{itemize}[nosep]
    \item A \textbf{consensus-driven multi-stage diagnostic interface} that visualizes consensus across explanations and evaluation metrics, enabling users to attribute interpretability failures to their source.
    \item A \textbf{three-phase semi-automatic labeling workflow} that combines active seeding with human judgment, facilitating the propagation of labels from a small seed set to the full feature set.
    \item A case study with three domain experts that demonstrates the system's effectiveness in diagnosing interpretability failures across 16,384 Sparse Autoencoder features.
\end{itemize}

\section{Background and Related Work}
\label{sec:related_work}

Throughout this paper, we use \emph{interpretation process} to refer to the process of understanding model internals, including the automated interpretation process previously described, and \emph{explanation} to refer to a natural-language description of a specific interpretable unit.

Our work focuses on diagnosing the automated interpretation process on the features extracted by Sparse Autoencoders (SAEs). This section introduces the technical background on SAEs and their automated interpretation process (\cref{subsec:sae}), reviews related visual analytics systems (\cref{subsec:va}) and interactive labeling approaches (\cref{subsec:al}), and identifies the challenges that motivate our work (\cref{subsec:challenge}).

\subsection{Sparse Autoencoders and Feature Interpretation}
\label{subsec:sae}
SAEs \cite{bricken2023monosemanticity, cunningham2023sparse} are the most widely adopted framework for decomposing LLM internals into interpretable units \cite{he2025saif, o2024disentangling, farrell2024applying, wu2025interpreting, gallifant2025sparse, nanda2024progress, ferrando2025do, muhamed2025saes}.
An SAE learns to decompose a model activation ${x \in \mathbb{R}^{d_{\text{model}}}}$ from a given layer into a sparse, high-dimensional representation $f \in \mathbb{R}^{d_{\text{SAE}}}$ (where $d_{\text{SAE}}\gg d_{\text{model}}$). We refer to each unit $f_i$ of this representation as a \emph{feature}. Specifically, the encoder applies a linear projection followed by a nonlinearity: 
$f=\text{ReLU}(W_{\text{enc}}x+b_{\text{enc}})$.
The ReLU activation, combined with a sparsity-inducing objective, ensures that only a small subset of features (typically $<1\%$) have non-zero coefficients per input. The decoder then reconstructs the original activation as a linear combination of these active features: $\hat{x}=W_{\text{dec}}f+b_{\text{dec}}$.
The SAE is trained end-to-end by minimizing a loss that balances reconstruction fidelity with sparsity.

This architecture offers several properties critical for interpretability. The sparsity constraint encourages each feature $f_i$ to represent a single concept, while the coefficient value of $f_i$, or \textbf{activation strength}, can be used to quantify the concept's strength in a given input. 
Furthermore, the columns of $W_{\text{dec}}$ define feature directions in activation space, enabling analysis of structural relationships via geometric proximity. For example, high cosine similarity between decoder vectors suggests possible feature splitting \cite{bricken2023monosemanticity}, a phenomenon where a single concept is distributed across multiple features.

Not all feature splits produce meaningful sub-concepts. Chanin et al. \cite{chanin2024aisfor} identified feature absorption, where a feature appears to track a concept but has arbitrary exceptions on which another feature activates instead. We extend this notion to \textbf{incoherent splitting}, where the boundary between two split features is arbitrary rather than conceptual. For example, two features might activate on arbitrary subsets of mathematics-related terms without discernible distinction. Because they appear monosemantic in isolation, per-feature evaluation fails to detect the problem. Such incoherent splits can undermine reliability and compromise downstream applications that assume a single concept is fully captured by a feature.

Since SAE features are learned in an unsupervised manner, mapping these features to explanations requires an additional post-hoc generation step. 
A straightforward approach is to examine \textbf{activating examples}, or text segments that activate a feature, to manually deduce the common semantic or linguistic pattern.
For example, a feature activating on terms such as ``integral'', ``derivative'', and ``algebra'' captures the semantic concept of mathematics, whereas one activating on ``freind'', ``adress'', or ``recieve'' indicates a linguistic pattern related to typos.

Manual inspection is infeasible at the typical SAE scale (16,384-262,144 features). Researchers therefore rely on surrogate explanation generators, or \textbf{explainers}, which are LLMs prompted to infer common patterns from feature-level evidence.
Bills et al. \cite{bills2023language} introduced LLM-based automatically generated explanations for individual neurons, and subsequent work extended this to SAE features with input-based \cite{foote2023n2g, gao2025scaling, paulo2025automatically} and output-based generation variants \cite{gur2025enhancing, wu2025interpreting}.

As explainers are prone to generating unfaithful explanations, prior work has introduced \textbf{interpretability metrics} \cite{paulo2025automatically, gao2025scaling, puri2025fade} to audit explanation faithfulness. These metrics also rely on LLM-based evaluators, referred to as \textbf{scorers}, which operate orthogonally to explainers.
A separate line of work assesses decomposition quality directly through supervised proxy tasks \cite{karvonen2025saebench, chanin2024aisfor, paulo2025automatically, chaudhary2024evaluating, farrell2024applying, marks2025sparse}.
\cref{fig:autoproc-diagpipe}-A summarizes this automated interpretation process.

\subsection{Visual Analytics for LLM Interpretability}
\label{subsec:va}
Visual analytics for LLM interpretability has primarily focused on helping users understand what models encode by visualizing the model internals \cite{hohman2019visual, barsoveanu2024visualizing}.
Early work visualizes individual components such as recurrent hidden states \cite{strobelt2017lstmvis, strobelt2018seq} and attention patterns \cite{vig-2019-multiscale, derose2020attention, wang2021dodrio, yeh2023attentionviz}. Subsequent systems broaden the scope to overall model behavior: interactive tutorials walk users through transformer architectures \cite{gao2023transforlearn, cho2025transformer}, and LM-TT \cite{tufanov2024lm} traces prediction-relevant information flow across layers.

Recent systems use automatically generated explanations as semantic entry points for exploring internal representations. NeuronautLLM \cite{woodman2024exploring} visualizes neuron activations and generated explanations to support prompt-driven neuron-level inspection in LLMs.
ConceptViz \cite{li2025conceptviz} brings concept-driven exploration, previously applied to vision models \cite{zhao2021human, huang2022conceptexplainer, wang2023drava}, to SAE features to find and validate concept correspondences. These systems are effective for targeted exploration because they start from a prompt or concept as a semantic anchor; however, they are unsuited for discovering unknown interpretability failures in an uncurated feature collection, especially when the explanations themselves are unreliable.
Neuronpedia \cite{neuronpedia} scales up inspection with per-feature dashboards that pair explanations and interpretability scores with activating examples, along with explanation search, but offers little guidance on which of thousands of features to inspect for failures.

Beyond interpretability, VISTA \cite{xuan2025vista} combines automated checks with human review to correct model-generated labels, and AttributionScanner \cite{xuan2025attributionscanner} helps experts find data slices on which a model behaves problematically. DECODE extends this line of human-in-the-loop validation from model outputs to the explanations of model internals.

To address these gaps, DECODE focuses on diagnosing the automated interpretation process itself, verifying whether it provides reliable evidence for what each feature captures.
It guides users to features worth inspecting, supports pairwise comparison for structural failures, summarizes consensus and metric evidence, and helps users attribute failures to either learned features or generated explanations.

\subsection{Active Learning and Visual Interactive Labeling}
\label{subsec:al}
Active learning (AL) addresses instance selection by having a model select instances that maximally improve classifier performance \cite{settles2009active, fu2013survey}. Recent work shows that optimal strategy depends on the labeling budget \cite{hacohen2022active}, prompting approaches that transition between strategies \cite{hacohen2023select, doucet2024bridging}.

Visual interactive labeling (VIAL) \cite{bernard2018vial} is a conceptual process model that integrates AL-driven and user-driven selection, motivated by evidence that user-driven selection can match or outperform pure AL, especially early in labeling \cite{bernard2017comparing}. Building on VIAL, Chegini et al. \cite{chegini2019interactive} suggest labels for high-confidence instances while directing users to uncertain instances via linked visualizations. Grimmeisen et al. \cite{grimmeisen2023visgil} encode AL utility measures into visual cues to guide instance selection. cVIL \cite{matt2025scalable} introduces a class-centric paradigm with property-measure-based selection and batch labeling. 
Beyond instance selection, prior work addresses post-hoc correction of mislabeled data \cite{xiang2019interactive, bauerle2020classifier} and visualizes how label distributions and classifier behavior evolve across labeling iterations \cite{beckmann2022interactive}.

DECODE adapts the VIAL process and builds on budget-dependent strategies \cite{hacohen2022active, hacohen2023select, doucet2024bridging} but differs in two respects. First, it addresses a setting where no ground truth exists: whether a feature explanation is faithful requires human assessment informed by multi-faceted evidence. Second, instead of transitioning between strategies on a fixed schedule \cite{doucet2024bridging} or by validation accuracy \cite{hacohen2023select}, DECODE lets AL recommend instances while users decide when to transition and verify predictions where classifiers disagree before propagation.

\begin{figure*}[t]
    \centering
    \includegraphics[width=0.95\textwidth]{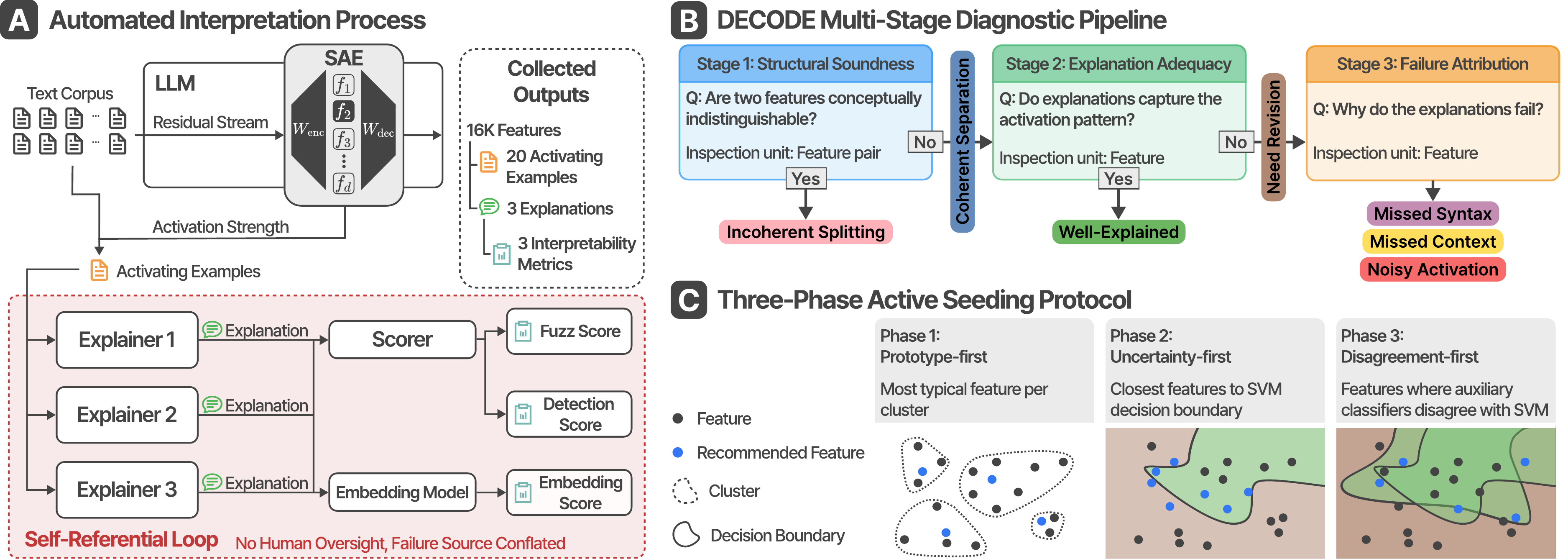}
    \caption{
    Overview of DECODE. 
    (A)~Automated interpretation process extracts SAE features, generates explanations with three explainers, and evaluates them with three metrics. The red dashed border highlights the self-referential loop.
    (B)~Three-stage diagnostic pipeline sequentially assigns features into five diagnostic categories, passing unresolved features to the next stage via intermediate categories.
    (C)~Three-phase active seeding protocol recommends features based on representativeness, decision boundary proximity, and classifier ensemble disagreement.
    }
    \label{fig:autoproc-diagpipe}
    \vspace{-3mm}
    
\end{figure*}

\subsection{Challenges}
\label{subsec:challenge}
From the SAE interpretability and visual analytics literature reviewed above, we identify three challenges that impede human-in-the-loop diagnosis of interpretability failures. We focus on prior work that involves the automated interpretation process, including SAE-based feature extraction, LLM-based explanation generation, metric-based explanation evaluation, and visual inspection of SAE features.

\textbf{C1. Lack of Ground-Truth Reference.}
SAE features lack ground-truth explanations, leaving no objective standard against which to validate outputs of the automated interpretation process. Prior work has introduced interpretability metrics to evaluate explanation faithfulness without ground truth \cite{paulo2025automatically, gao2025scaling, puri2025fade}, but they remain surrogate signals rather than external references.
Different interpretability metrics assess different aspects of faithfulness and can disagree on the same explanation \cite{paulo2025automatically, puri2025fade}. Likewise, different explainers encode different priors and linguistic biases, often producing contradictory explanations for the same feature.
As a result, users are left without reliable guidance on which explanation to trust or which metric to prioritize.

\textbf{C2. Limited Scalability of Human Oversight.}
The scalability constraint that necessitated LLM-based automation also limits human oversight. SAEs typically learn 16,384 to 262,144 features, each associated with dozens of activating examples, multiple candidate explanations from different explainers, and a set of interpretability metrics per explanation.
Prior SAE studies therefore often rely on small sampled subsets for manual inspection, typically around a hundred features \cite{paulo2025automatically, bricken2023monosemanticity}. Visual interfaces make features easier to inspect \cite{li2025conceptviz, neuronpedia}, but human oversight still requires deciding where to focus attention and how to extend local judgments beyond individually examined features, both of which per-feature dashboards such as Neuronpedia leave to users.

\textbf{C3. Obscured Failure Attribution.}
The automated interpretation process consists of multiple dependent modules. The SAE decomposes model activations into features, the explainer generates an explanation from activating examples, and interpretability metrics evaluate the explanation. 
However, interpretability scores are sensitive to upstream decomposition quality. A low score conflates SAE-driven failures with explainer-driven failures, leaving users unable to attribute the failure to its source.
While supervised proxy tasks offer an independent check on decomposition quality \cite{karvonen2025saebench, chanin2024aisfor, chaudhary2024evaluating, farrell2024applying, marks2025sparse}, their limited coverage leaves failure attribution unresolved for the majority of features.

\section{The DECODE System}

\begin{table}[t]
\centering
\caption{Terminal diagnostic categories used in DECODE.}
\vspace{-3mm}
\label{tab:categories}
\setlength{\tabcolsep}{2pt}
\setlength{\defaultaddspace}{2pt}  %
\footnotesize
\begin{tabular}{p{0.22\linewidth} p{0.74\linewidth}}
\toprule
\textbf{Category} & \textbf{Description} \\
\midrule
\emph{Incoherent}\newline\emph{Splitting}
& An SAE-driven failure where two features divide a single concept without a clear conceptual boundary. \\
\addlinespace
\emph{Well-Explained}
& A feature whose generated explanations faithfully capture the shared pattern across its activating examples. \\
\addlinespace
\emph{Missed Syntax}
& An explainer-driven failure where the explanations omit a shared surface-level pattern, such as common subwords or morphological affixes. \\
\addlinespace
\emph{Missed Context}
& An explainer-driven failure where the explanations omit a shared contextual pattern, such as topic, domain, or tone.\\
\addlinespace
\emph{Noisy Activation}
& An SAE-driven failure where the feature activates on heterogeneous examples with no coherent pattern. \\
\bottomrule
\end{tabular}
\vspace{-4mm}
\end{table}

\subsection{Design Rationale}
Our DECODE (Diagnosing Explanations via COnsensus-Driven Evaluation) system is designed to address the aforementioned challenges, following the design rationales outlined below.

\textbf{DR1. Enable Diagnosis of Consensus Across Explainers and Metrics.}
To address the absence of ground-truth explanations (\textbf{C1}), we adopt an approach analogous to qualitative research, where interpretations are produced and cross-validated by multiple coders, and inter-coder consensus serves as a proxy for reliability.

We bring this consensus-based analogy to our setting in two ways. First, we treat each explainer as an independent coder and assume that explanations on which multiple explainers agree are more likely to be reliable (consensus among explainers). Second, since explainer consensus may reflect shared biases and individual metrics are insufficient to reliably capture explanation faithfulness, we complement the ensemble of explainers with an ensemble of interpretability metrics as an independent validation indicator (consensus among scorers).

\textbf{DR2. Provide Semi-Automatic Labeling via Active Seeding.}
To address the scalability limitation of manual inspection (\textbf{C2}), we reframe open-ended interpretability assessments into a structured classification where each feature is assigned one of five \textbf{diagnostic categories} (\cref{tab:categories}).
We derive these categories by separating \emph{Well-Explained} features and partitioning interpretability failures by source (SAE-driven vs. explainer-driven) and granularity. 
This categorization directly informs concrete actions: retaining reliable features for downstream tasks, retraining the SAE to resolve decomposition issues, partially regenerating explanations via targeted refined prompts, or filtering noise.
However, implementing such corrective loops is outside the scope of our work.

\begin{figure}[]
    \centering
    \includegraphics[width=0.95\columnwidth]{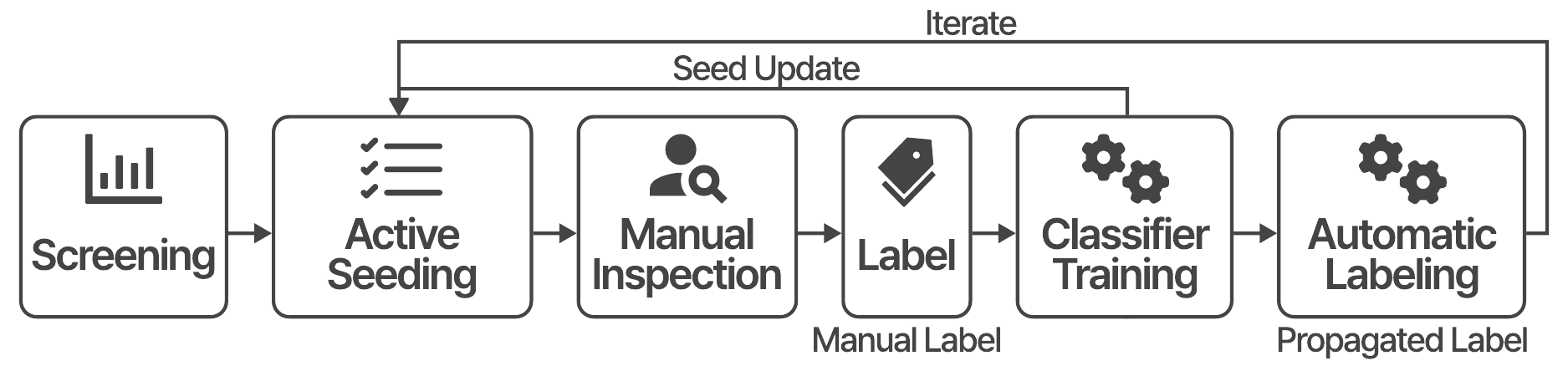}
    \caption{The iterative labeling workflow within each diagnostic stage. Users screen, inspect, and label features in a loop where each label retrains the classifier for automatic labeling.}
    \label{fig:semiauto}
    \vspace{-3mm}
\end{figure}

Manually classifying each feature into diagnostic categories remains a scalability challenge. To address this, we employ a semi-automatic labeling approach (\cref{fig:semiauto}). Under the assumption that features with similar profiles share the same diagnostic category, a classifier trained on a small manually labeled seed set can automatically label the remaining features. To maximize labeling efficiency, we design a three-phase active seeding protocol that adapts its sampling criteria to the classifier's maturity, which is further detailed in \cref{subsec:stage1}.

\textbf{DR3. Support Multi-Stage Failure Diagnosis.}
To enable failure attribution (\textbf{C3}) and reduce cognitive burden, we decompose the diagnostic pipeline into three stages, with each depending on the outcome of the previous stage (\cref{fig:autoproc-diagpipe}-B).
\textbf{Structural Soundness} first filters similar feature pairs that split one concept arbitrarily as \emph{Incoherent Splitting}, because feature-level analysis cannot detect upstream structural issues. \textbf{Explanation Adequacy} identifies \emph{Well-Explained} features from unnecessary failure attribution. Features with unfaithful explanations proceed to \textbf{Failure Attribution}, where each failure is attributed to its source.

Decomposition also reduces cognitive burden by limiting each stage to a single diagnostic question. Evaluating multiple interacting criteria simultaneously would exceed human working memory limits \cite{halford1998processing, sweller2010element}. A staged approach facilitates consistent judgments through focused sequential decisions, each supported by tailored coordinated views.

\subsection{User Tasks}
\label{subsec:task}

Our target users are expert interpretability researchers who inspect SAE features prior to downstream analyses such as safety auditing. DECODE supports users already familiar with SAE phenomena (e.g., feature splitting, noisy activations) who need to identify these issues across a large feature set. We derived our user tasks from established SAE evaluation protocols and case studies \cite{bricken2023monosemanticity, templeton2024scaling, chanin2024aisfor, puri2025fade} that rely on small-scale manual inspection and mapped these practices into our three diagnostic stages (\textbf{DR3}).

\textbf{T1. Identify Incoherent Feature Splitting.}
This task addresses incoherent splitting, an SAE-driven structural interpretability failure. Because splitting is a relational property, this task operates on pairs of features.
Prior work has performed this task through manual inspection, albeit at a limited scale. Templeton et al. \cite{templeton2024scaling} manually examined activating examples of geometrically neighboring features across SAE widths to verify a meaningful split. Chanin et al. \cite{chanin2024aisfor} compared activation patterns across features in the same SAE to threshold problematic features.
Following established practice, users compare activating examples across feature pairs to judge whether the shared patterns correspond to meaningful sub-concepts or reflect incoherent splitting.

\textbf{T2. Assess Explanation Faithfulness.}
This task identifies features whose explanations do not faithfully represent the underlying activation pattern. 
Bricken et al. \cite{bricken2023monosemanticity} demonstrated this via a manual evaluation protocol where evaluators judge explanation faithfulness against activating examples. Following this protocol, users first discover the underlying pattern from activating examples, then judge whether the generated explanations from multiple explainers match the discovered pattern. Users further cross-validate these judgments against interpretability scores, which test whether the explanations are specific enough to distinguish activating examples from non-activating examples.

\textbf{T3. Attribute Failure Source of Misinterpreted Features.}
This task targets features identified as misinterpreted in (\textbf{T2}) and attributes each interpretability failure to its source.
Prior work has recognized the need for such attribution; for example, FADE \cite{puri2025fade} distinguishes description-level misalignment from feature-level defects across four interpretability metrics.
Building on these taxonomies, users classify each feature into one of three diagnostic categories (\textbf{DR2}, \textbf{DR3}): \emph{Missed Syntax} or \emph{Missed Context} for explainer-driven failures, and \emph{Noisy Activation} for SAE-driven failures.

\subsection{Data Processing}
\label{subsec:dataprocess}
We instantiate our pipeline on Gemma-2-9b, an open-weight language model widely studied in the interpretability community \cite{rajamanoharan2024jumping, lieberum2024gemma, gallifant2025sparse}. We use the SAE trained on residual stream activations at layer 31 ($\frac{3}{4}$ of the way), following the standard approach established in Gao et al. \cite{gao2025scaling}, which decomposes the 3,584-dimensional residual stream ($d_{\text{model}}=3,584$) into 16,384 sparse features ($d_{\text{SAE}}=16,384$). All SAE weights and per-feature activating data are sourced from Neuronpedia \cite{neuronpedia}. Each of the $d_{\text{SAE}}$ features is associated with 20 activating examples ($N_{\text{act}}=20$), three independently generated explanations ($N_{\text{exp}}=3$) with each explanation having three interpretability scores ($N_{\text{metr}}=3$), and an explainer consensus score ($S_{\text{consensus}}$).

\textbf{Activating Examples.}
We use stratified sampling over activation strength \cite{paulo2025automatically} rather than selecting only top-activated examples to capture the full range of activation behavior. For each feature, we extract activating examples from Neuronpedia \cite{neuronpedia}, which provides token-level activation strength across a text corpus. We divide activating examples into four activation-strength quantiles and sample five examples from each, yielding $N_{\text{act}}$ per feature. Each example retains its token-level activation strengths, which serve as both visual encoding in the interface and weights for embedding computation. 

\textbf{Explanations.}
We employ three independent explainers ($N_{\text{exp}}=3$) to establish explainer consensus: Llama 3.1 70B \cite{grattafiori2024llama} (Llama), GPT-4o mini \cite{hurst2024gpt} (GPT), and Gemini 2.5 Flash-Lite \cite{comanici2025gemini} (Gemini).
For Llama, deployed locally, we feed $N_{\text{act}}$ stratified activating examples as described above for explanation generation, adapting the approach by Paulo et al. \cite{paulo2025automatically}.
GPT and Gemini explanations are sourced from Neuronpedia, with different input configurations. Implementation details for all three explainers are provided in the supplementary materials.

\textbf{Interpretability Metrics.}
We evaluate each explanation using $N_{\text{metr}}$ metrics adopted from Paulo et al. \cite{paulo2025automatically}. The activating examples help users inspect what pattern a feature captures, but they do not test whether an explanation is specific to those activations. We therefore use complementary metrics measuring how well an explanation $z$ discriminates activating examples $\mathcal{A}$ from randomly sampled non-activating examples $\mathcal{N}$ over $n$ examples. Each interpretability score ranges from 0 to 1, with 0.5 as a random baseline.

\textit{Fuzz score} evaluates token-level discrimination: within $n$ examples from $\mathcal{A}$, the scorer distinguishes correctly marked activating tokens from randomly marked non-activating tokens given $z$.
\textit{Detection score} evaluates context-level discrimination: given $z$, a scorer classifies $n$ examples from $\mathcal{A} \cup \mathcal{N}$ as activating or non-activating.
\textit{Embedding score} evaluates semantic alignment without a scorer: it reports the AUROC of cosine similarities between the sentence embedding of $z$ and those of $n$ examples in $\mathcal{A} \cup \mathcal{N}$.
Detailed configurations are in the supplementary materials.

We use Llama as the scorer. Interpretability scores are computed per feature-explainer pair, yielding nine scores per feature ($N_{\text{metr}} \times N_\text{exp}=9$). Using the same model as both explainer and scorer may introduce self-preference bias \cite{panickssery2024llm}; however, all three metrics frame evaluation as classification tasks rather than open-ended quality judgments, thereby mitigating subjective bias.
We concatenate the three scores per explanation to establish consensus among scorers (\textbf{DR1}), indicating whether an explanation satisfies independent evaluation criteria.

\textbf{Explainer Consensus.}
To quantify consensus among explainers (\textbf{DR1}), we first decompose each explanation into constituent phrases by dependency-based clause splitting. All phrases across $N_{\text{exp}}$ explainers for a given feature $i$ are then embedded and clustered via agglomerative clustering with average linkage. The clustering threshold is selected to maximize mean inter-cluster similarity; details are provided in supplementary materials.
The resulting clusters group semantically similar phrases. Clusters with contributions from multiple explainers represent genuine convergence, with the medoid phrase serving as the consensus description. Conversely, single-explainer clusters and singletons indicate divergence among explainers.

To ensure equal contribution regardless of verbosity, each phrase $j$ from explainer $e$ (producing $p_e$ phrases) is assigned weight $w_{e,j} = 1/p_e$. Given $K$ clusters $\{C_1, \dots, C_k, \dots,  C_K\}$, where $|E_k^{(i)}|$ denotes the number of explainers in cluster $k$, we define:
{
\setlength{\abovedisplayskip}{3pt}
\setlength{\belowdisplayskip}{3pt}
\begin{align}
    S_{k}^{(i)} &= \frac{|E_k^{(i)}|-1}{N_{\text{exp}}-1}\sum_{(e,j) \in C_k^{(i)}} w_{e,j}^{(i)} \\
    S_{\text{consensus}}^{(i)} &= \frac{1}{N_{\text{exp}}}\sum_{k=1}^{K} S_k^{(i)}
\end{align}
}
A high cluster score $S_{k}^{(i)}$ indicates descriptions on which explainers broadly agree. The feature-level score $S_{\text{consensus}}^{(i)} \in [0,1]$ aggregates and normalizes across clusters, reaching 1 when every phrase is clustered and each cluster involves all $N_{\text{exp}}$ explainers.

\subsection{User Interface}
\begin{figure*}[t]
    \centering
    \includegraphics[width=0.95\textwidth]{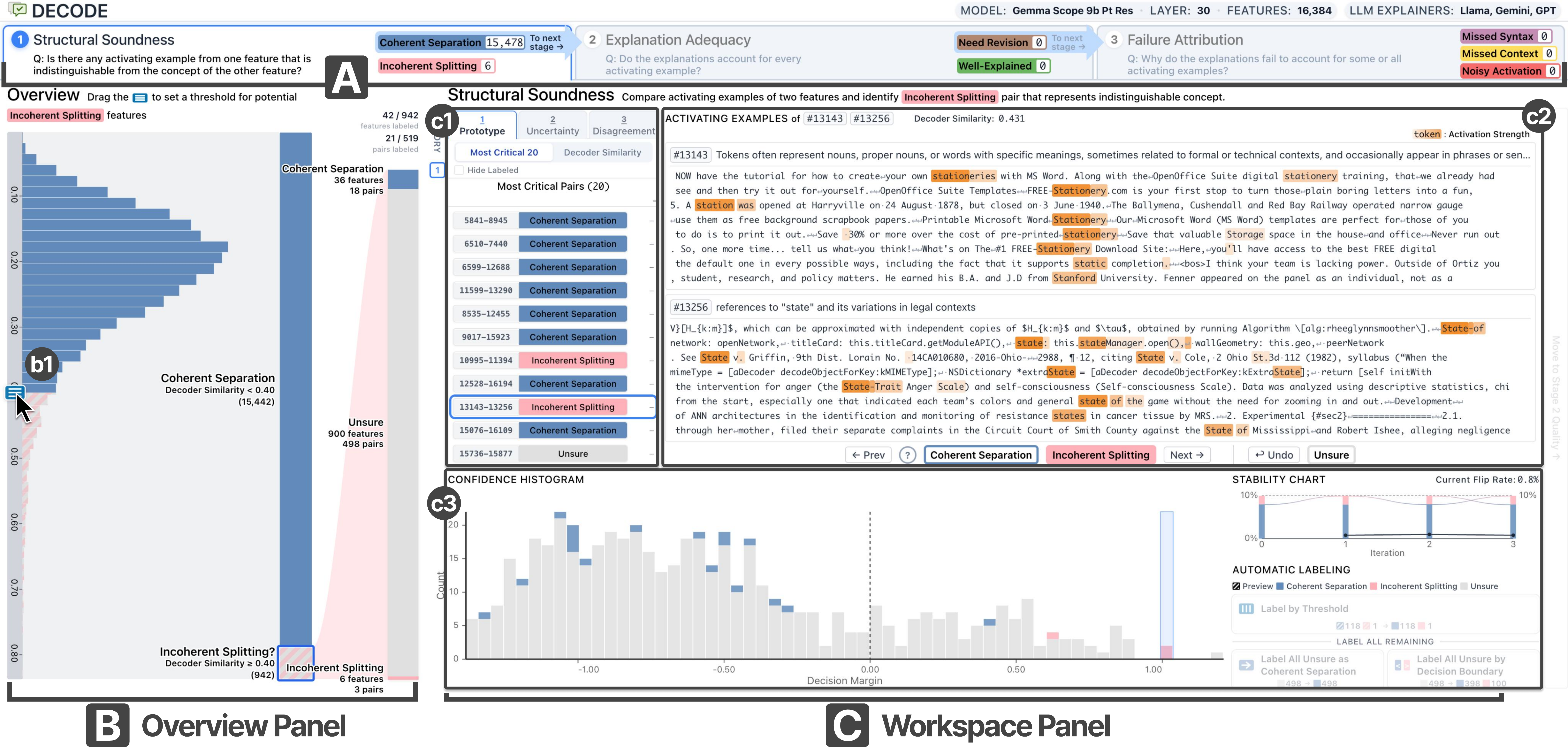}
    \caption{Stage~1 interface. 
    (A)~Stage tabs indicate diagnostic progress (Stage 1 active). 
    (B)~Overview Panel displays a decoder similarity histogram. A draggable threshold at 0.40 separates inspection candidates (above) from features assigned \emph{Coherent Separation} (below).
    (C)~Workspace Panel: (c1)~Seeding List ranks candidate pairs, showing 20 representatives in prototype-first phase; (c2)~Inspection View juxtaposes feature 13143 and 13256 both activating on tokens starting with ``sta''; (c3)~Labeling View displays confidence histogram and stability chart after three training iterations.
    }
    \label{fig:stage1}
    \vspace{-3mm}
    
\end{figure*}

The goal of the DECODE system{\footnote{Our code will be released under the MIT License upon acceptance.}} is to assign each SAE feature one of five diagnostic categories (\textbf{DR2}). To make this feasible at scale, the interface decomposes diagnosis into three stages and employs a human-in-the-loop semi-automatic labeling at every stage (\textbf{DR3}).

\cref{fig:stage1} presents the interface. The interface is organized into three tabs, one per diagnostic stage (\cref{fig:stage1}-A). Across all stages, the \textbf{Overview Panel} (\cref{fig:stage1}-B) on the left provides an overview of the classification process. 
The \textbf{Workspace Panel} (\cref{fig:stage1}-C) on the right contains three coordinated views that adapt to each stage. First, the \textbf{Seeding List} (\cref{fig:stage1}-c1) shows high-priority features for manual inspection via active seeding (\textbf{DR2}). Next, the \textbf{Inspection View} (\cref{fig:stage1}-c2) presents stage-specific detail for the selected feature to enable manual inspection. Finally, the \textbf{Labeling View} (\cref{fig:stage1}-c3) enables automatic labeling using a classifier trained on previously labeled features.
To reduce the learning curve, DECODE provides inline tooltips for technical terms and a stage-specific labeling guide next to the labeling buttons, summarizing each category with decision criteria and examples.

\begin{figure*}[t]
    \centering
    \includegraphics[width=0.95\textwidth]{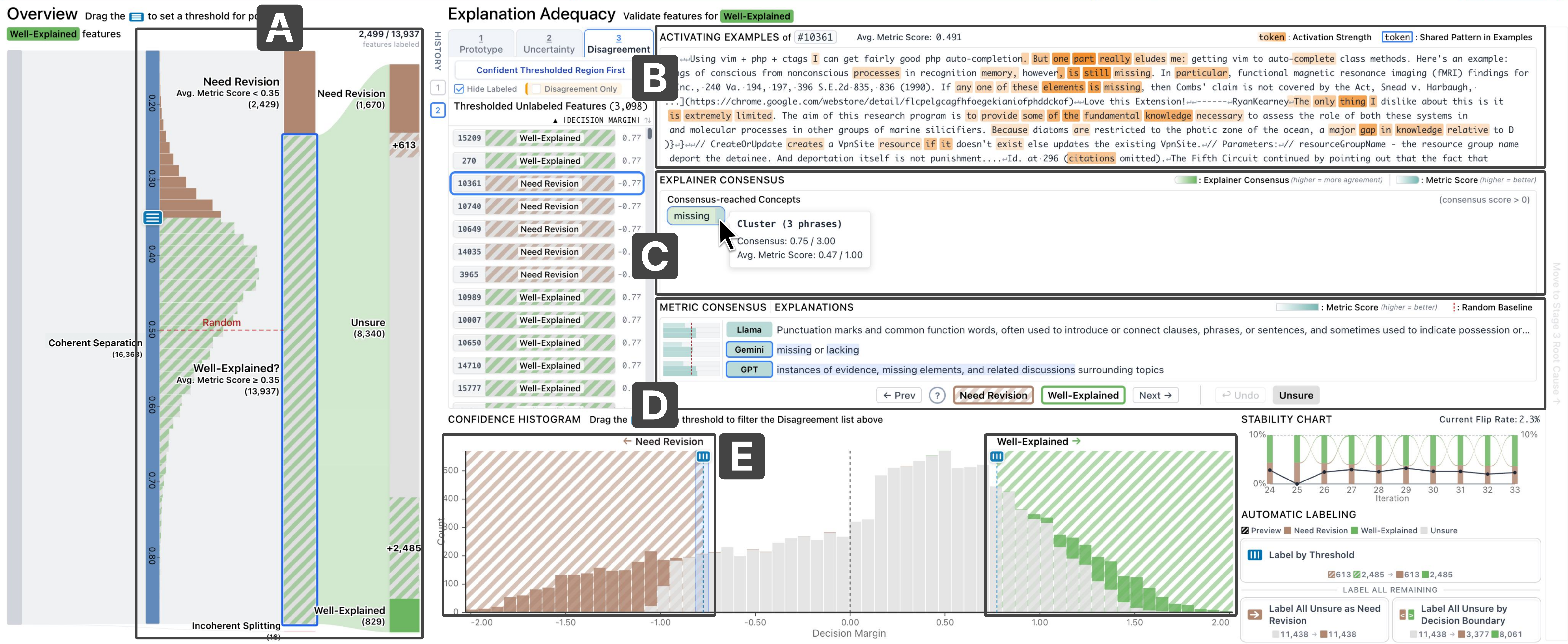}
    \caption{Stage~2 interface. 
    (A)~Overview Panel screens features by average interpretability score as in Stage 1.
    Feature 10361 is chosen for inspection: 
    (B)~activating examples with absence-related tokens highlighted;
    (C)~the user hovers over the explainer consensus cluster with medoid ``missing'';
    (D)~constituent phrases of the hovered cluster highlighted in blue, with per-explainer interpretability scores showing fuzz and detection above random baseline;
    (E)~confidence histogram with draggable thresholds defining the automatic labeling region.
    }
    \label{fig:stage2}
        \vspace{-3mm}
\end{figure*}

\begin{figure*}[t]
    \centering
    \includegraphics[width=0.95\textwidth]{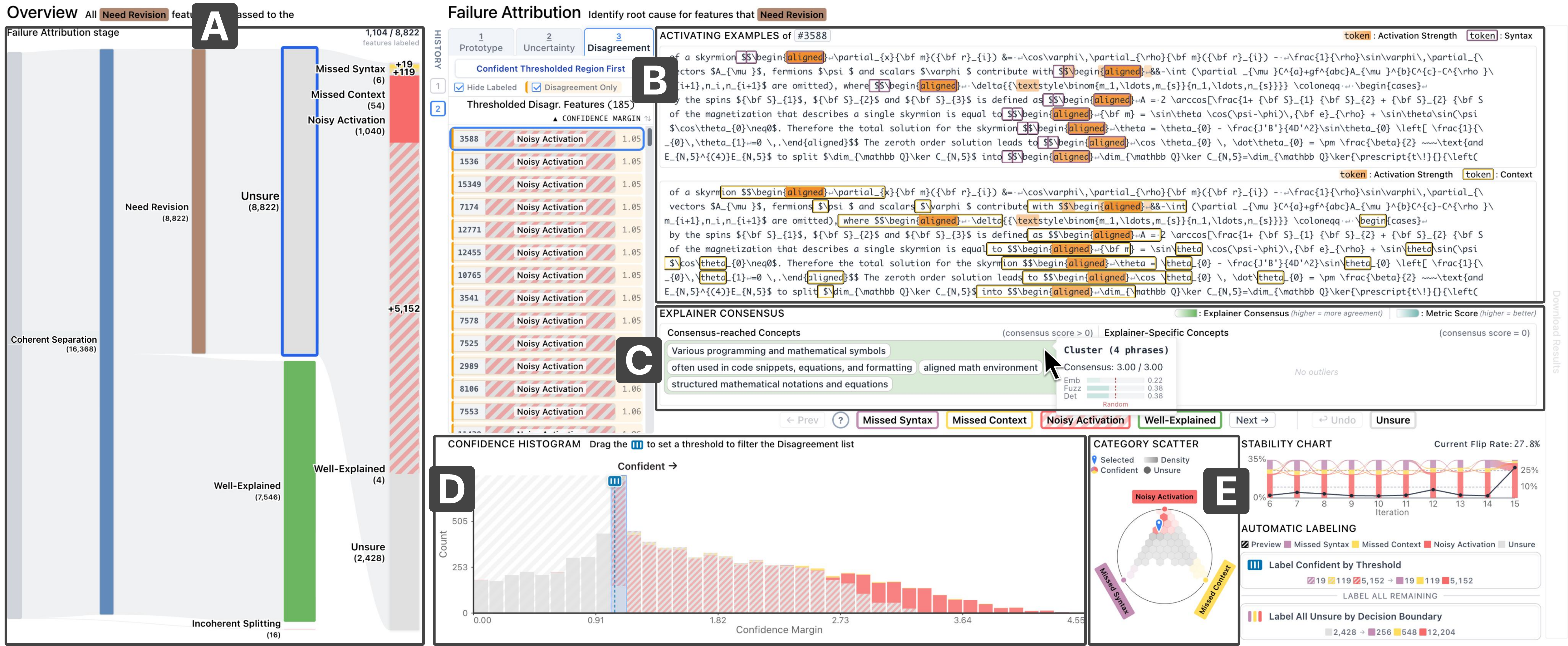}
    \caption{Stage~3 interface for feature 3588. 
    (A)~Overview Panel shows all \emph{Need Revision} features passed from Stage 2 without threshold-based screening. 
    The Inspection View presents: 
    (B)~two activating-example panels, with syntax-pattern borders on ``\$\$'' and ``aligned'' (top) and context-pattern borders on semantically similar phrases and distinctive recurring tokens such as ``\$\textbackslash'' and ``\textbackslash theta'' (bottom);
    (C)~expanded explainer consensus clusters listing all constituent phrases with per-explainer scores on hover.
    The Labeling View adapts to three categories: 
    (D)~confidence histogram with tentative labels predominantly \emph{Noisy Activation}; 
    (E)~category scatter showing distribution skewed toward \emph{Noisy Activation}.
    }
    \label{fig:stage3}
    \vspace{-3mm}
    
\end{figure*}

The Overview Panel (\cref{fig:stage2}-A and \cref{fig:stage3}-A) summarizes diagnostic progress using a Sankey diagram.
Each vertical layer of the diagram corresponds to a stage, with the leftmost layer representing the unlabeled SAE features. 
The Sankey diagram extends to the right; the rightmost layer (i.e., vertically stacked bars) shows the features that are categorized in the current stage. 

We adopt a Sankey diagram for three reasons. First, Sankey diagrams have been widely used to visualize the progressive splitting and transition of entities, which naturally aligns with our diagnostic pipeline, where features are hierarchically assigned to diagnostic categories across stages. Second, the height of each node encodes the number of features, providing a quantitative overview of feature distribution at each stage. Third, since our diagnostic pipeline consists of a fixed number of stages and categories, the visualization maintains its scalability and interpretability even as the number of features increases.

Throughout the interface, we consistently represent a feature as a rectangular glyph annotated with its ID.
The color of the glyph indicates the diagnostic category or initial \emph{Unsure} state, and the hatching pattern distinguishes confirmed assignments (solid) from tentative assignments suggested by the system (stripe).

\subsubsection{Stage 1: Structural Soundness}
\label{subsec:stage1}
The goal of Stage 1 is to identify features with \emph{Incoherent Splitting} (\textbf{T1}). As splitting is a relational property, diagnosis operates on feature pairs (i.e., two features that may represent similar concepts).
Pairs labeled as \emph{Incoherent Splitting} are excluded from further analysis.
The remaining features are assigned the intermediate category \emph{Coherent Separation} and proceed to subsequent stages. This category indicates only that no incoherent splitting was identified under the examined pairwise comparisons, rather than monosemanticity.

Identifying incoherent splitting requires finding semantically similar features. High cosine similarity between decoder vectors indicates potential feature splitting \cite{templeton2024scaling} (\cref{subsec:sae}).
Computing pairwise decoder similarities across all $N$ features yields $O(N^2)$ candidate pairs, which is intractable at typical SAE scales.
To address this, the Overview Panel (\cref{fig:stage1}-B) supports screening through an interactive histogram of decoder similarity. Users adjust a threshold via a draggable handle (\cref{fig:stage1}-b1; default: 0.40); features falling below this threshold are automatically assigned \emph{Coherent Separation}, and the rest are paired and presented in the Seeding List for manual inspection (see supplementary material for pair generation details).
Lowering the threshold includes more candidate features with weaker geometric proximity, prioritizing recall over precision to avoid missing borderline cases.

For semi-automatic labeling, we represent each pair as an 8-dimensional vector of pair activation statistics and decoder vector similarity (see supplementary material) and train a Support Vector Machine (SVM) to generalize human judgments to unlabeled pairs.
We choose SVM as the primary classifier for three reasons. First, its fast inference enables sub-second feedback loops for interactive analysis. Second, its max-margin objective generalizes well from few labels \cite{Vapnik2000} and mitigates the cold-start problem. Third, it enables efficient uncertainty estimation based on the distance to the decision boundary.

The Seeding List (\cref{fig:stage1}-c1) ranks candidate pairs for manual inspection according to our three-phase seeding protocol (\textbf{DR2}, \cref{fig:autoproc-diagpipe}-C). Each phase appears as a selection tab, between which users can switch.

In the \textit{prototype-first} phase, the list presents an initial seed set to mitigate the cold-start problem. With few labels, uncertainty cannot be reliably estimated, and labeling typical examples is more effective \cite{hacohen2022active}. We therefore adopt TypiClust, clustering the pair representation space into 20 groups and selecting the most typical pair from each cluster, where typicality is measured by density within the cluster.

After all pairs in the seed set are labeled, users can switch to the \textit{uncertainty-first} phase. As labels accumulate, querying uncertain examples becomes more effective \cite{hacohen2022active, doucet2024bridging}; the list therefore ranks pairs by ascending absolute distance from the decision boundary, surfacing ambiguous cases where splitting is neither clearly present nor absent.

When the classifier predictions stabilize, users activate the final \textit{disagreement-first} phase, which supplements the SVM with two auxiliary classifiers (a multi-layer perceptron and a random forest) in a query-by-committee approach \cite{seung1992qbc}. We choose them because their inductive biases differ from SVM's max-margin boundary. Pairs where the SVM is confident but the committee majority disagrees are surfaced for verification before automatic labeling, flagging potential confident misclassifications that uncertainty-first ranking misses.
To preserve the influence of manually verified pairs, automatically labeled pairs are assigned a lower sample weight of 0.2 during classifier retraining \cite{matt2025scalable}.

For each selected pair, the Inspection View (\cref{fig:stage1}-c2) vertically juxtaposes the activating examples of each paired feature. To provide an initial interpretive anchor, the explanation with the highest average interpretability score is displayed above each feature's activating examples. We sample two activating examples from each of the four quantiles for each feature to capture the complete activation behavior.
To facilitate comparison between the examples (\textbf{T1}), we encode per-token activation strength using background opacity, following established conventions \cite{neuronpedia}.
Users assign the pair \emph{Incoherent Splitting}, \emph{Coherent Separation}, or \emph{Unsure} with three labeling buttons.
The \emph{Unsure} option allows users to defer judgment on ambiguous pairs; these pairs are excluded from classifier training and resolved via automatic labeling at the end of the stage.

The Labeling View (\cref{fig:stage1}-c3) supports a reliable transition from manual inspection to automatic labeling by visualizing the classifier's predictions and prioritizing confident regions for automatic labeling. The classifier is first trained once each category has at least three labeled pairs and retrains as users label pairs. If the seed set falls short of this minimum, users find additional candidates by sorting pairs by decoder similarity.
The transition requires users to answer two questions: how reliably the current classifier categorizes remaining unlabeled pairs, and when additional labeling yields diminishing returns.

To address the former question, the confidence histogram (\cref{fig:stage1}-c3, left) visualizes the signed distance from the decision boundary.
Pairs at the extremes are confidently classified, while those near zero are ambiguous. Each bar is stacked to separate labeled and unlabeled pairs, letting users verify whether their manual assignments align with the classifier's predictions.
The stability chart (\cref{fig:stage1}-c3, top right) addresses the latter question by complementing the snapshot with how predictions flip across labeling iterations.
A declining flip rate approaching zero indicates that additional labels yield diminishing reclassification. A line chart conveys the trend of the flip rate, while links between per-iteration stacked bars of predicted categories show the direction of flips, which the rate alone cannot reveal.

When the flip rate converges and users activate the disagreement-first tab, two draggable threshold handles appear on the confidence histogram to designate high-confidence regions at both extremes for automatic labeling.
Pairs in these regions are ranked in the Seeding List for disagreement verification.

Upon verification, users apply automatic labeling through three buttons (\cref{fig:stage1}-c3, bottom right). The \emph{Apply by Threshold} button assigns categories to all pairs within the confidence thresholds. Users can iteratively adjust thresholds, verify candidates, and re-apply to expand labeling coverage.
To avoid prolonged iteration, two fallback buttons address remaining unlabeled pairs.
The \emph{Apply by Boundary} assigns categories by the current decision boundary without threshold filtering.
The \emph{Apply Safe Default} button assigns \emph{Coherent Separation} to all remaining pairs, accepting lower recall of \emph{Incoherent Splitting} in exchange for ensuring no feature is prematurely excluded from subsequent diagnosis.

Throughout labeling, a stacked bar extending from the Sankey diagram shows the remaining \emph{Unsure} features, and checkpoints record each automatic batch assignment so that users can revert it.

\subsubsection{Stage 2: Explanation Adequacy}
The goal of Stage 2 is to assess whether generated explanations faithfully represent the underlying activation pattern (\textbf{T2}, \cref{fig:stage2}). While Stage 1 operates on feature pairs, Stage 2 shifts to individual features. Each \emph{Coherent Separation} feature from the previous stage is labeled \emph{Well-Explained} if the explanation matches the activation pattern across all its activating examples and distinctly separates activating from non-activating examples. Features that fail either condition are assigned the intermediate category \emph{Need Revision} for further failure attribution.

The Overview Panel replaces decoder similarity with the average interpretability score as its screening criterion (\cref{fig:stage2}-A). Features scoring below a user-specified threshold are automatically assigned \emph{Need Revision} and excluded from manual inspection, following the same screening logic as Stage 1. The default threshold is set to 0.35, filtering out low-scoring features while roughly balancing the two categories. The Seeding List applies the same three-phase active seeding protocol, now ranking individual features represented as an 11-dimensional vector of interpretability score statistics, consensus measures, and activation statistics, detailed in the supplementary material.

Users label each feature as \emph{Well-Explained}, \emph{Need Revision}, or \emph{Unsure}.
The Inspection View facilitates this judgment through three vertically stacked components, forming a progressive workflow from raw evidence to consensus among explainers and scorers.
Users first examine the feature's activating examples, where
recurring $n$-grams are outlined in blue to help users identify the underlying pattern (\cref{fig:stage2}-B). Identifying the pattern independently before seeing the generated explanations helps users notice patterns that the explainers may have missed.

Users then match their discovered pattern against explainer consensus clusters computed in \cref{subsec:dataprocess}. Each cluster shows its medoid phrase, and background opacity encodes the cluster-level consensus score (\cref{fig:stage2}-C). Hovering over a cluster highlights the corresponding source phrases in each explanation displayed below. A match between the discovered pattern and high-scoring clusters supports a \emph{Well-Explained} assignment, while misalignment signals \emph{Need Revision} (\textbf{T2}).

Users finally cross-validate with each explainer's interpretability scores (\textbf{T2}).
For each explainer, three scores are displayed as a horizontal bar group with 0.5 marked as a random baseline (\cref{fig:stage2}-D). Agreement between manual assessment and the ensemble of scores reinforces confidence in judgment. Discrepancies, particularly when a score falls below a random baseline (i.e., worse than random discrimination), prompt users to revisit the activating examples under the metric's evaluation criteria. For example, a low fuzz score alongside a high detection score suggests that the explainer captured the overall context but overlooked specific token-level patterns.

The Labeling View retains the same confidence histogram, stability chart, and automatic labeling controls, now placing \emph{Well-Explained} and \emph{Need Revision} on opposite sides of the decision boundary.
The \emph{Apply Safe Default} button assigns \emph{Need Revision}, ensuring ambiguous features proceed to failure attribution in the next stage.

\subsubsection{Stage 3: Failure Attribution}
The goal of Stage 3 is to attribute each interpretability failure to its source for features assigned \emph{Need Revision} in Stage 2 (\textbf{T3}, \cref{fig:stage3}). Users assign one of the diagnostic categories \emph{Noisy Activation}, \emph{Missed Syntax}, and \emph{Missed Context}, with an additional \emph{Well-Explained} option to override the judgment if closer inspection reveals the explanation faithful. As the final stage, all features receive a diagnostic category.

Unlike previous stages, no single screening criterion reliably separates the three diagnostic categories prior to inspection. Threshold-based filtering is omitted from the Overview Panel, and all \emph{Need Revision} features directly enter the Seeding List. The feature representation remains the same 11-dimensional vector as in Stage 2. To handle three categories, the SVM extends to a One-vs-One configuration, training a binary classifier for each category pair and aggregating their decisions. The classifier is first trained once each category has at least two labeled features, and the prototype-first phase selects 30 seeds instead of 20.

Since this stage assumes the explanations are unfaithful, the diagnostic question shifts from whether they are faithful to what specifically the explainers missed.
The Inspection View presents two identical sets of activating examples that pre-annotate surface-level and semantic patterns, respectively (\cref{fig:stage3}-B), helping users quickly identify present patterns before examining explainer consensus clusters.
Accordingly, the explainer consensus clusters from Stage 2 are reconstructed to serve as a concept inventory: each cluster expands to list all constituent phrases rather than showing only the medoid (\cref{fig:stage3}-C) so that users can quickly survey which patterns the explainers captured. Hovering reveals cluster-aggregated per-explainer interpretability scores.

To determine the failure source, users cross-reference this inventory against the activating examples, identifying patterns visible in the examples but absent from the clusters.
If activating examples lack discernible patterns and cluster consensus scores are low, the failure indicates \emph{Noisy Activation}.
If a surface-level pattern in the upper panel or a semantic pattern in the lower panel is absent from the explainer consensus clusters, it indicates \emph{Missed Syntax} or \emph{Missed Context}, respectively.

The Labeling View adapts to the three-category setting. The confidence histogram encodes the Best-vs-Second-Best margin (\cref{fig:stage3}-D), where a small margin indicates high uncertainty. Users adjust a threshold that partitions features into an unsure region and a confident region, which enter the Seeding List during the uncertainty-first and disagreement-first phases, respectively.
The margin, however, does not indicate which categories a feature lies between, and per-category confidence histograms would show each feature in three separate plots.
We therefore adapt RadViz \cite{hoffman1997radviz} into a category scatter that places each feature at the confidence-weighted mean of three points on a circle, one per category (\cref{fig:stage3}-E), which for three categories is equivalent to a ternary plot.
Automatic labeling provides \emph{Apply by Threshold} and \emph{Apply by Boundary} buttons. Unlike previous stages, no safe default exists at this terminal stage.

\section{Case Study}
\label{sec:evaluation}

We evaluate DECODE through case studies with three domain experts.
As SAE-based interpretability is a nascent field and the diagnostic task requires substantial domain knowledge, we opted for in-depth case studies with experts over a controlled user study.
The task was to diagnose the automated interpretation process of a given SAE by assigning each feature one of the five diagnostic categories (\cref{tab:categories}).
Experts worked through the three-stage pipeline, manually inspecting features surfaced by active seeding and propagating their judgments to the full feature set through automatic labeling.
We used the full 16,384 SAE features from Gemma-2-9B (layer 31), and all feature IDs reported in this section are publicly available on Neuronpedia \cite{neuronpedia}.

All three experts had foundational knowledge of SAEs and research experience in model interpretability. 
P1, a researcher holding a Master's degree with 1.5 years of experience in interpretability, led SAE-based interpretability research for Dense Passage Retrieval models.
P2, a PhD student with three years of experience in interpretability and alignment, holds a top-tier publication in the area.
P3, a PhD student with one year of experience in interpretability, published research on using SAE features to identify and ablate toxicity.

Each session lasted three hours, including breaks, and consisted of four parts. The session began with a 10-minute pre-interview to assess each expert's prior experience with SAE features and the automated interpretation process. A 30-minute tutorial then covered the necessary background (e.g., SAE explanation generation and interpretability metrics), diagnostic categories with representative examples, and a guided walkthrough of the interface. Each expert independently performed the full three-stage pipeline over approximately 120 minutes. The session concluded with a 20-minute semi-structured post-interview gathering insights on diagnostic strategies and feedback on the system. Experts were compensated at 70 USD per hour.

\subsection{Labeling Effort}
\label{subsec:result}

All three experts completed the full three-stage pipeline within a single session, spending 127, 112, and 132 minutes on 138, 167, and 168 labels (\cref{tab:effort}).
We refer to these labels that experts assigned after inspecting individual features as \textbf{manual labels}, and those the classifier assigns to the remaining features as \textbf{propagated labels}.
Of the manually labeled features, 88, 85, and 106 received a terminal diagnostic category (\cref{tab:distribution}); the remainder fell into the intermediate categories (\emph{Coherent Separation} and \emph{Need Revision}).
Time per unit rose monotonically across stages for every expert, consistent with experts reporting the three-way attribution as the most demanding judgment (\cref{subsec:feedback}).

\emph{Incoherent Splitting} was rare, covering 57, 18, and 42 features, and \emph{Well-Explained} accounted for 45 to 51\% of the feature set. Among the failures, \emph{Noisy Activation} was dominant for all three, indicating that most failures were SAE-driven rather than explainer-driven. The experts diverged mainly between \emph{Missed Syntax} and \emph{Missed Context}.

\begin{table}[t]
\caption{Labeling effort per stage. A label denotes one manually labeled unit: a feature pair in Stage 1, a feature in Stages 2 and 3.}
\vspace{-2mm}
\label{tab:effort}
\centering
\footnotesize
\setlength{\tabcolsep}{4pt}
\setlength{\abovetopsep}{0pt}
\setlength{\aboverulesep}{0.2ex}
\setlength{\belowrulesep}{0.32ex}
\renewcommand{\arraystretch}{0.92}
\begin{tabular}{l rrr rrr rrr}
\toprule
& \multicolumn{3}{c}{Stage 1} & \multicolumn{3}{c}{Stage 2} & \multicolumn{3}{c}{Stage 3} \\
\cmidrule(lr){2-4}\cmidrule(lr){5-7}\cmidrule(lr){8-10}
& Labels & Min & /unit & Labels & Min & /unit & Labels & Min & /unit \\
\midrule
P1 & 59 & 49 & 0.83 & 45 & 42 & 0.93 & 34 & 36 & 1.06 \\
P2 & 79 & 42 & 0.53 & 36 & 27 & 0.75 & 52 & 43 & 0.83 \\
P3 & 59 & 28 & 0.47 & 48 & 38 & 0.79 & 61 & 66 & 1.08 \\
\bottomrule
\end{tabular}
\vspace{-2mm}
\end{table}

\begin{table}[t]
\centering
\caption{Category distribution per expert. \emph{Well-Explained} aggregates assignments made at Stage 2 and at Stage 3. Each cell shows manual feature count with the final total after automatic labeling in parentheses.}
\vspace{-2mm}
\label{tab:distribution}
\setlength{\tabcolsep}{2pt}
\setlength{\aboverulesep}{0.2ex}
\setlength{\belowrulesep}{0.32ex}\footnotesize
\begin{tabular}{lcccccc}
\toprule
 & \makecell{Incoherent\\Splitting} & \makecell{Well-\\Explained} & \makecell{Missed\\Syntax} & \makecell{Missed\\Context} & \makecell{Noisy\\Activation} & Total \\
\midrule
P1 & 26 (57)    & 41 (8,403)  &  3 (70)    &  2 (731)   & 16 (7,123)  &  88 (16,384) \\
P2 & 16 (18)    & 40 (7,567)  &  6 (1,187)  &  3 (243)   & 20 (7,369)  &  85 (16,384) \\
P3 & 14 (42) & 61 (7,319) &  4 (820)   &  2 (590)   & 25 (7,613) & 106 (16,384) \\
\bottomrule
\vspace{-6mm}
\end{tabular}
\end{table}

\begin{table}[t]
\centering
\caption{End-to-end macro-F1 on the reference set. Every condition assigns all five diagnostic categories to all 16,384 features, combining manual labels with propagated labels from automatic labeling.}
\vspace{-2mm}
\label{tab:accuracy}
\setlength{\tabcolsep}{2.5pt}
\setlength{\aboverulesep}{0.2ex}
\setlength{\belowrulesep}{0.32ex}
\renewcommand{\arraystretch}{0.95}
\footnotesize
\begin{tabular}{lcccccccc}
\toprule
 & Score-only & \multicolumn{3}{c}{Individual expert} & \multicolumn{3}{c}{Expert pair} & Group of \\
\cmidrule(lr){3-5}\cmidrule(lr){6-8}
 & baseline & P1 & P2 & P3 & P1-P2 & P1-P3 & P2-P3 & 3 experts \\
\midrule
F1 & 0.36 & 0.40 & 0.37 & 0.41 & 0.53 & 0.54 & 0.50 & 0.62 \\
\bottomrule
\end{tabular}
\vspace{-4mm}
\end{table}

\subsection{Label Quality}
\label{subsec:acc}
To examine how label quality differs between human oversight and automation alone, we constructed a reference set from the 233 features that at least one expert manually labeled with a terminal category.
Of the 233 features, 200 carry one manual label, 20 carry two, and 13 carry three.
The reference set was labeled independently by an LLM (Claude Opus 4.8) and one author, each judging from the same interface evidence. The LLM generated a rationale for each label, which the author consulted to adjudicate a final label (see supplementary materials).
Expert labels remained blinded.
Building the reference set took approximately eight hours, allowing more time per feature than the expert sessions permitted.
The reference set serves as a comparison standard across conditions that reflects the author's interpretation of the diagnostic categories rather than independent ground truth.
We report macro-F1 rather than accuracy, as \emph{Missed Syntax} and \emph{Missed Context} account for only 16 and 11 of the 233 features and accuracy is dominated by the other majority categories.

To assess the label quality with respect to the degree of human involvement, we compare four conditions.
The \textbf{score-only baseline} replaces human judgment with DECODE's default screening thresholds, applying them as labeling rules rather than as filters for manual inspection: it assigns \emph{Incoherent Splitting} and \emph{Well-Explained} to features above the Stage 1 and 2 thresholds, and the category matching each feature's highest z-normalized metrics at Stage 3 (see supplementary materials for threshold tuning analysis and corresponding metrics).
Each \textbf{individual expert} is one expert's session, in which manual labels on a seeded subset propagate to the remaining features.
The \textbf{expert pair} and the \textbf{group of experts} aggregate two and all three experts, preferring manual labels wherever they exist and resolving overlap by majority. All unbroken ties are counted as incorrect.

The three individual experts reached 0.40, 0.37, and 0.41 macro-F1, marginally above the 0.36 of the score-only baseline (\cref{tab:accuracy}), but the aggregated score obscures two opposite error trade-offs.
The baseline terminated most features as \emph{Incoherent Splitting} or \emph{Well-Explained} at Stages 1 and 2, reaching 77\% and 84\% recall respectively, but collapsed on failure attribution: it recovered \emph{Noisy Activation} at 10\% recall, assigning 58\% of these features to \emph{Well-Explained}.
The experts showed the reverse: 23-44\% and 55-75\% recall on the first two categories but 67-87\% on \emph{Noisy Activation}.
The experts recovered SAE-driven failures that the baseline classified as valid.

An individual expert score combines manual and propagated labels, so we scored the two separately.
Manual labels alone yielded 0.60, 0.65, and 0.63 macro-F1, while propagated labels yielded 0.29, 0.23, and 0.25, suggesting that automatic labeling accounts for the modest individual expert scores.
Because active seeding selects ambiguous features and each expert's classifier propagates a criterion fitted to a distinct subset of them, the reference set is ultimately skewed toward features that are challenging for automatic labeling.

Aggregating expert judgments improved macro-F1 over every individual session: 0.53, 0.54, and 0.50 for the pairs and 0.62 for the group (\cref{tab:accuracy}). This gain stems from both coverage and majority voting. Coverage expanded manual labels from 88, 85, and 106 features for the individuals to 157, 169, and 173 for the pairs, and to the full set for the group, replacing the propagated labels with manual ones; for the group, it changed 259 labels summed over the three sessions for a net gain of 158 correct labels.
Majority voting applied to the 33 features with two or three manual labels; for the group, it changed 21 labels for a net gain of 9.
Voting helped because expert disagreements were systematic: the experts reached moderate to substantial agreement \cite{landis1977measurement} (Cohen’s $\kappa$ .481–.663) and differed mainly in strictness between \emph{Well-Explained} and \emph{Noisy Activation} (see supplementary materials).

\subsection{Observed Labeling Behavior}
\label{subsec:strat}

We report the patterns common to all experts, illustrated with representative features, and the points where their strategies diverged.

\textbf{Pairwise activation comparison exposed failures that per-feature evaluation would have missed.}
For feature pair 13143 and 13256 (\cref{fig:stage1}), P1 found in the activating examples that feature 13143 activated on words starting with ``sta'' (``station,'' ``static,'' ``Stanford''), while the paired feature 13256 activated only on ``state.'' Feature 13143 appeared to cover every word starting with the prefix ``sta'' except ``state,'' an arbitrary rather than conceptual boundary, so P1 labeled the pair \emph{Incoherent Splitting}. Per-feature explanations were faithful (e.g., ``references to `state'''), and the failure appeared only when the two features were read side by side.

\textbf{Explainer consensus clusters anchored interpretation when the activating examples were ambiguous.}
For feature 582, P1 could not identify a concept from the activating examples alone, and drew a candidate from the explainer consensus cluster ``careful examination of records.'' Re-examining the examples, P1 found several mismatches and labeled the feature \emph{Need Revision}.
The clusters also anchored P2, who read them before the examples. For feature 10361 (\cref{fig:stage2}), P2 read the cluster ``missing,'' searched for the pattern, found few matches, and assigned \emph{Need Revision}. In Stage 3, closer inspection revealed the pattern expressed through paraphrases such as ``citation omitted,'' and P2 overrode the label to \emph{Well-Explained}. P2 overrode two more features (15101, 5026) likewise, an anchoring effect we revisit in \cref{subsec:lessons}.

\textbf{Interpretability scores played a limited role in decision-making.}
P1 disregarded them as reflecting the ambiguity of the activating examples rather than the faithfulness of the explanation, and P3 reported little influence on any decision. Only P2 consulted them to confirm \emph{Well-Explained} assignments, but retained the initial judgment in every case: scores above the random baseline confirmed it, and below-baseline scores triggered a re-examination that revealed no new pattern.

\subsection{Expert Feedback}
\label{subsec:feedback}
\textbf{Perceived Benefits of the System.}
From a practical standpoint, P1, who had previously conducted human evaluation of SAE features using spreadsheets, noted that ``having a tool like this would have been much more convenient.''
Experts highlighted the responsiveness of the interactive labeling loop. P1 found the interactive model refinement ``well-pipelined for quick feedback.''
P3, who had never used a feature exploration tool, reported that ``the interface felt dense at first but posed no difficulty once the stage structure became familiar.''

At the component level, all experts found the stability chart important and effective for monitoring classifier maturity.
P1 confirmed that ``the effect of incremental training was clearly visible.''
P3 noted that ``in most cases it settled down as I kept tagging,'' treating a sudden jump as a cue to reconsider whether the label just assigned was consistent with earlier ones.
In Stage 3, P2 used category scatter to monitor category distribution, noting that ``seeing the distribution made me think I should look for more \emph{Missed Context} features.''

\textbf{Suggestions for Improvement.}
All experts found Stage 3 the most challenging due to the finer-grained distinctions required. Decomposing the three-way decision into a hierarchical binary sequence (i.e., first SAE-driven vs. explainer-driven, then surface-level vs. semantic) could reduce this burden. P3 added that ``surface-level patterns were easier to overlook than semantic ones'', suggesting that syntactic cues warrant stronger visual emphasis.
All experts noted that the high degree of freedom in confidence threshold selection increased cognitive load unrelated to the diagnostic task. Providing sensible defaults while preserving manual override could balance guidance with flexibility. Beyond committee-disagreed features, P1 proposed the system ``show a random sample of confidence-thresholded features'' for post-labeling verification.
P3 wanted to test an interpretation directly, entering custom text to see whether the feature actually activates, resolving questions that fixed activating examples cannot.

\section{Discussion}
Our case studies demonstrate three benefits of reintroducing human oversight that automated interpretability pipelines do not provide.
First, human judgment recovered failures that automation certified as sound (\cref{subsec:acc}).
Second, the staged pipeline attributed each failure to a specific component, as every stage inspects at the level where its target failure is visible (\cref{subsec:strat}). 
Third, the diagnosis yielded actionable insights into the pipeline. All three experts independently identified \emph{Noisy Activation} as the predominant failure mode, attributing most errors to the decomposition step rather than the explainers (\cref{subsec:result}). This guides corrective action directly: retraining the SAE addresses these failures, while refining explainer prompts would leave them untouched.

DECODE requires interpretable units with token-level activations and generated explanations, and Stage 1 further relies on decoder vectors. It extends beyond SAEs to per-layer and cross-layer transcoders \cite{ameisen2025circuit}, while only Stages 2 and 3 apply to individual neurons \cite{bills2023language}.

\subsection{Lessons Learned}
\label{subsec:lessons}
Beyond these benefits, our design and evaluation surfaced three lessons that revisit the challenges of the lack of ground truth (C1) and limited human oversight (C2) in human-in-the-loop diagnostic workflows.

\textbf{Leverage explainer consensus as a proxy, but safeguard it against anchoring bias (C1)}.
With the lack of ground truth, all experts used explainer consensus clusters as their primary tool for understanding feature semantics: for clear features, consensus alone was sufficient to reach a confident judgment; for ambiguous features, explainer consensus clusters served as an interpretive anchor that redirected experts' initial hypotheses (\cref{subsec:strat}).
However, consulting consensus before forming an independent interpretation can introduce anchoring bias. For instance, P2, who reviewed the explainer consensus clusters prior to examining the activating examples, subsequently overrode three labels due to this anchoring effect.
P1 and P3, who consistently formed a hypothesis from the activating examples before consulting clusters, did not exhibit this bias.
Systems providing multi-source consensus should enforce an evidence-first workflow so that consensus serves as validation rather than a starting point.

\textbf{Aggregate judgments from multiple experts rather than relying on a single one (C1).}
A single expert's judgment cannot be externally validated either, so consensus among experts serves as a proxy for reliability. Aggregating expert judgments improved label quality over any individual (\cref{subsec:acc}): coverage replaced propagated labels with manual ones, and majority voting resolved disagreements among experts.
Since disagreements concentrate on boundary cases, full multi-expert review is unnecessary; systems can instead route only low-confidence or boundary features to additional experts and record the resolved judgments in a shared codebook, as in qualitative coding practice.

\textbf{Stabilize category boundaries before propagating labels through automation (C2).}
Human oversight scales through automatic labeling, but this is also where reliability degrades: boundary decisions are made once, in isolation, and then propagated to the full feature set, with no persistent reference to keep them consistent. The cost is highest for the rarest failure sources, whose boundaries rested on as few as two labels.
The shared codebook introduced above can serve as this missing reference.
Future iterations could maintain a codebook of anchor features per diagnostic category, initialized from the tutorial.

\subsection{Limitations}

\textbf{Label Quality.}
The group of experts reached only 0.62 macro-F1 (\cref{subsec:acc}). Propagated labels scored lower (0.23–0.29) as each classifier extends one expert’s criterion from a few ambiguous features. This label quality still suffices to identify the predominant failure source, on which all three experts agreed. It also supports restricting downstream analyses to \emph{Well-Explained} features more reliably than the score-only baseline, which assigned most \emph{Noisy Activation} features to \emph{Well-Explained}. Fine-grained attribution of individual features, however, remains unreliable, particularly between \emph{Missed Syntax} and \emph{Missed Context}, where the experts' category distributions diverged most (\cref{tab:distribution}).

\textbf{Cold Start and Class Imbalance.}
Learning from few labels poses two problems for the classifier. The first is cold start: the classifier must begin from only a few labels. The SVM classifier and prototype-first seeding mitigate this, but the seed set covers the representation space only coarsely, and confident misclassifications in uncovered regions may not be detected in the disagreement-first phase.
The second is class imbalance: some categories, such as \emph{Missed Syntax} and \emph{Missed Context}, remain scarce however many labels accumulate. The minimum number of labels per category ensures that every category is represented, not that it is represented adequately.

\textbf{Scope of Diagnostic Framework.}
Our five-category diagnostic taxonomy may not exhaustively cover all interpretability failures. For instance, the pipeline assumes that an explanation is legible and therefore does not assess readability separately. We address only incoherent splitting among structural failures, leaving feature hedging \cite{chanin2025feature} outside the framework, and polysemantic features or output-based features \cite{arad2025saes} may warrant a distinct category from \emph{Noisy Activation}.

\textbf{Threats to External Validity.}
We instantiated DECODE on a single SAE configuration (Gemma-2-9b, layer 31, 16,384 features). Different model families, layer positions, and SAE widths may produce features with different distributional characteristics.
We also fixed the three explainers, the three interpretability metrics, the default screening thresholds, and the SVM classifier, and did not measure how diagnoses vary under alternative choices.
The experts, all familiar with SAE phenomena, completed the pipeline in about two hours after a 30-minute tutorial but still found Stage 3 demanding; applicability to less experienced users remains untested.
Finally, the classifier assumes that features with similar profiles belong to the same diagnostic category, which fails when distinct interpretability failures manifest identically.

\section{Conclusion}
\label{sec:conclusion}

We present DECODE, a visual analytics system that reintroduces human oversight into the automated interpretation process. DECODE uses consensus among explainers and metrics as a proxy for reliability, decomposes diagnosis into three stages to attribute failures, and scales human judgment through semi-automatic labeling. Case studies with three domain experts showed that it identifies the predominant source of interpretability failures.

\bibliographystyle{ieeetr}
\bibliography{refs}

\end{document}